\documentclass[10pt,twocolumn,letterpaper]{article}

\usepackage{cvpr}
\usepackage{times}
\usepackage{epsfig}
\usepackage{graphicx}
\usepackage{amsmath}
\usepackage{amssymb}
\usepackage{bm}
\usepackage{makecell}
\usepackage{arydshln}
\usepackage{multirow}
\usepackage{gensymb}
\usepackage{float}

\usepackage[pagebackref=true,breaklinks=true,letterpaper=true,colorlinks,bookmarks=false]{hyperref}

\cvprfinalcopy 

\def\cvprPaperID{1536} 

\ifcvprfinal\fi
\begin{document}

\title{ASLFeat: Learning Local Features of Accurate Shape and Localization}

\author{Zixin Luo$^{1}$\hspace{0.7cm} Lei Zhou$^{1}$\hspace{0.7cm} Xuyang Bai$^{1}$\hspace{0.7cm} Hongkai Chen$^{1}$\hspace{0.7cm} Jiahui Zhang$^{2}$ \\ Yao Yao$^{1}$\hspace{0.7cm} Shiwei Li$^{3}$\hspace{0.7cm} Tian Fang$^{3}$\hspace{0.7cm} Long Quan$^{1}$ \\
\normalsize $^1$Hong Kong University of Science and Technology \\ \normalsize $^2$Tsinghua University \hspace{0.7cm} \normalsize $^3$Everest Innovation Technology \\
\tt\small\{zluoag,lzhouai,xbaiad,hchencf,yaoag,quan\}@cse.ust.hk \\
\tt\small jiahui-z15@mails.tsinghua.edu.cn\hspace{0.7cm} \{sli,fangtian\}@altizure.com}

\maketitle
\thispagestyle{empty}

\begin{abstract}
This work focuses on mitigating two limitations in the joint learning of local feature detectors and descriptors.
First, the ability to estimate the local shape (scale, orientation, etc.) of feature points is often neglected during dense feature extraction, while the shape-awareness is crucial to acquire stronger geometric invariance. Second, the localization accuracy of detected keypoints is not sufficient to reliably recover camera geometry, which has become the bottleneck in tasks such as 3D reconstruction. In this paper, we present ASLFeat, with three light-weight yet effective modifications to mitigate above issues. First, we resort to deformable convolutional networks to densely estimate and apply local transformation. Second, we take advantage of the inherent feature hierarchy to restore spatial resolution and low-level details for accurate keypoint localization. Finally, we use a peakiness measurement to relate feature responses and derive more indicative detection scores. The effect of each modification is thoroughly studied, and the evaluation is extensively conducted across a variety of practical scenarios. State-of-the-art results are reported that demonstrate the superiority of our methods. [\href{https://github.com/lzx551402/aslfeat}{code release}]
\end{abstract}

\section{Introduction}
Designing powerful local features is an essential basis for a broad range of computer vision tasks~\cite{schonberger2016structure, zhang2017distributed,zhou2018learning,sattler2012image,zhang2019learning,li2015dual,zhang2019learning}. During the past few years, the joint learning of local feature detectors and descriptors has gained increasing popularity, with promising results achieved in real applications. However, there are two limitations we consider that may have hinged further boost in performance: 1) the lack of shape-awareness of feature points for acquiring stronger geometric invariance, and 2) the lack of keypoint localization accuracy for solving camera geometry robustly.

Traditionally, the local shape is parameterized by hand-crafted scale/rotation estimation~\cite{lowe2004distinctive,rublee2011orb} or affine shape adaptation~\cite{mikolajczyk2002affine}, while more recently, data-driven approaches~\cite{moo2016learning,mishkin2018repeatability,yi2016lift} have emerged that build a separate network to regress the shape parameters, then transform the \emph{patch inputs} before feature descriptions. Due to the increasing prevalence of the joint learning with keypoint detectors~\cite{detone2018superpoint,ono2018lf,revaud2019r2d2,dusmanu2019d2,christiansen2019unsuperpoint}, recent research focus has shifted to frameworks that densely extract features from \emph{image inputs}, whereas no pre-defined keypoint is given and thus previous patch-wise shape estimation becomes inapplicable. 
As an alternative, LF-Net~\cite{ono2018lf} extracts dense features and transforms intermediate feature maps via Spatial Transformer Networks (STN)~\cite{jaderberg2015spatial}, whereas multiple forward passes are needed and only sparse predictions of shape parameters are practically feasible. In this view, there still lacks a solution that enables efficient local shape estimation in a dense prediction framework.

Besides, the localization accuracy of learned keypoints is still concerned in solving geometry-sensitive problems. For instance, LF-Net~\cite{ono2018lf} and D2-Net~\cite{dusmanu2019d2} empirically yield low precision in two-view matching or introduce large reprojection error in Structure-from-Motion (SfM) tasks, which in essence can be ascribed to the lack of spatial accuracy as the detections are derived from low-resolution feature maps (e.g., $1/4$ times the original size). To restore the spatial resolution, SuperPoint~\cite{detone2018superpoint} learns to upsample the feature maps with pixel-wise supervision from artificial points, while R2D2~\cite{revaud2019r2d2} employs dilated convolutions to maintain the spatial resolution but trades off excessive GPU computation and memory usage. Moreover, it is questionable that if the detections from the deepest layer are capable of identifying low-level structures (corners, edges, etc.) where  keypoints are often located. Although widely discussed in dense prediction tasks~\cite{ronneberger2015u,godard2017unsupervised,lin2017refinenet}, in our context, neither the keypoint localization accuracy, nor the low-level nature of keypoint detection has received adequate attention.

To mitigate above limitations, we present ASLFeat, with three light-weight yet effective modifications. First, we employ deformable convolutional networks (DCN)~\cite{dai2017deformable,zhu2019deformable} in the dense prediction framework, which allows for not only pixel-wise estimation of local transformation, but also progressive shape modelling by stacking multiple DCNs. Second, we leverage the inherent feature hierarchy, and propose a multi-level detection mechanism that restores not only the spatial resolution without extra learning weights, but also low-level details for accurate keypoint localization. Finally, we base our methods on an improved D2-Net~\cite{dusmanu2019d2} that is trained from scratch, and further propose a peakiness measurement for more selective keypoint detection.

Despite the key insights of above modifications being familiar,  we address their importance in our specific context, fully optimize the implementation in a non-trivial way, and thoroughly study the effect by comparing with different design choices. To summarize, we aim to provide answers to two critical questions: 1) what \emph{deformation parameterization} is needed for local descriptors (geometrically constrained~\cite{moo2016learning,mishkin2018repeatability,yi2016lift} or free-form modelling~\cite{dai2017deformable,zhu2019deformable}), 2) what \emph{feature fusion} is effective for keypoint detectors (multi-scale input~\cite{revaud2019r2d2,dusmanu2019d2}, in-network multi-scale inference~\cite{ono2018lf}, or multi-level fusion~\cite{ronneberger2015u}). Finally, we extensively evaluate our methods across various practical scenarios, including image matching~\cite{balntas2017hpatches,bian2019evaluation}, 3D reconstruction~\cite{schonberger2017comparative} and visual localization~\cite{sattler2012image}. We demonstrate drastic improvements upon the backbone architecture, D2-Net, and report state-of-the-art results on popular benchmarks.
\section{Related works}
Hand-crafted local features have been widely evaluated in~\cite{balntas2017hpatches,schonberger2017comparative}, we here focus mainly on the learning approaches.

\smallskip\noindent\textbf{Local shape estimation.}
Most existing descriptor learning methods~\cite{luo2018geodesc,luo2019contextdesc,mishchuk2017working,tian2019sosnet,tian2017l2,zhang2019learning} do not explicitly model the local shape, but rely on geometric data augmentation (scaling/rotational perturbation) or hand-crafted shape estimation (scale/rotation estimation~\cite{lowe2004distinctive,rublee2011orb}) to acquire geometric invariance. Instead,  OriNet~\cite{moo2016learning} and LIFT~\cite{yi2016lift} propose to learn a canonical orientation of feature points, AffNet~\cite{mishkin2018repeatability} predicts more affine parameters to improve the modelling power, and the log-polar representation~\cite{ebel2019beyond} is used to handle in particular scale changes. Despite the promising results, those methods are limited to take \emph{image patches} as input, and introduce a considerable amount of  computation since two independent networks are constructed for predicting patch shape and patch description separately. As an alternative, UCN~\cite{choy2016universal} and LF-Net~\cite{ono2018lf} takes \emph{images} as input and performs STN~\cite{jaderberg2015spatial} on intermediate features. In particular, LF-Net requires multiple forward passes to transform individual ``feature patch", and thus only prediction on sparse locations is practically applicable.

Meanwhile, the modelling of local shape has been shown crucial in image recognition tasks, which inspires works such as scale-adaptive convolution (SAC) for flexible-size dilations~\cite{zhang2017scale} and deformable convolution networks (DCN) for tunable grid sampling locations~\cite{dai2017deformable,zhu2019deformable}. In this paper, we adopt the similar idea in our context, and propose to equip DCN for dense local transformation prediction, of which the inference requires only a single forward pass and is thus of high efficiency.

\smallskip\noindent\textbf{Joint local feature learning.}
The joint learning of feature detectors and descriptors has received increasing attention, where a unified network is constructed to share most computations of the two tasks for fast inference. In terms of descriptor learning, the ranking loss~\cite{ono2018lf,dusmanu2019d2,detone2018superpoint,christiansen2019unsuperpoint,revaud2019r2d2} has been primarily used as a \emph{de-facto} standard. However, due to the difficulty of acquiring unbiased ground-truth data, no general consensus has yet been reached regarding an effective loss design for keypoint detector learning. For instance, LF-Net~\cite{ono2018lf} warps the detection map and minimizes the difference at selected pixels in two views, while SuperPoint~\cite{detone2018superpoint} operates a self-supervised paradigm with a bootstrap training on synthetic data and multi-round adaptations on real data. More recent R2D2~\cite{revaud2019r2d2} enforces grid-wise peakiness in conjunction with reliability prediction for descriptor, while UnsuperPoint~\cite{christiansen2019unsuperpoint} and Key.Net~\cite{laguna2019key} learn grid-wise offsets to localize keypoints.

By contrast, D2-Net~\cite{dusmanu2019d2} eschews learning extra weights for a keypoint detector, but hand-crafts a selection rule to derive keypoints from the same feature maps that are used for extracting feature descriptors. This design essentially couples the capability of the feature detector and descriptor, and results in a clean framework without complex heuristics in loss formulation. However, it is a known issue that D2-Net lacks of accuracy of keypoint localization, as keypoints are derived from low-resolution feature maps. In this paper, we base our methods on D2-Net, and mitigate above limitation by a light-weight modification that cheaply restores both the spatial resolution and low-level details.
\section{Methods}
\subsection{Prerequisites}
\label{sec:prer}
\begin{figure*}
    \centering
    \includegraphics[width=0.88\textwidth]{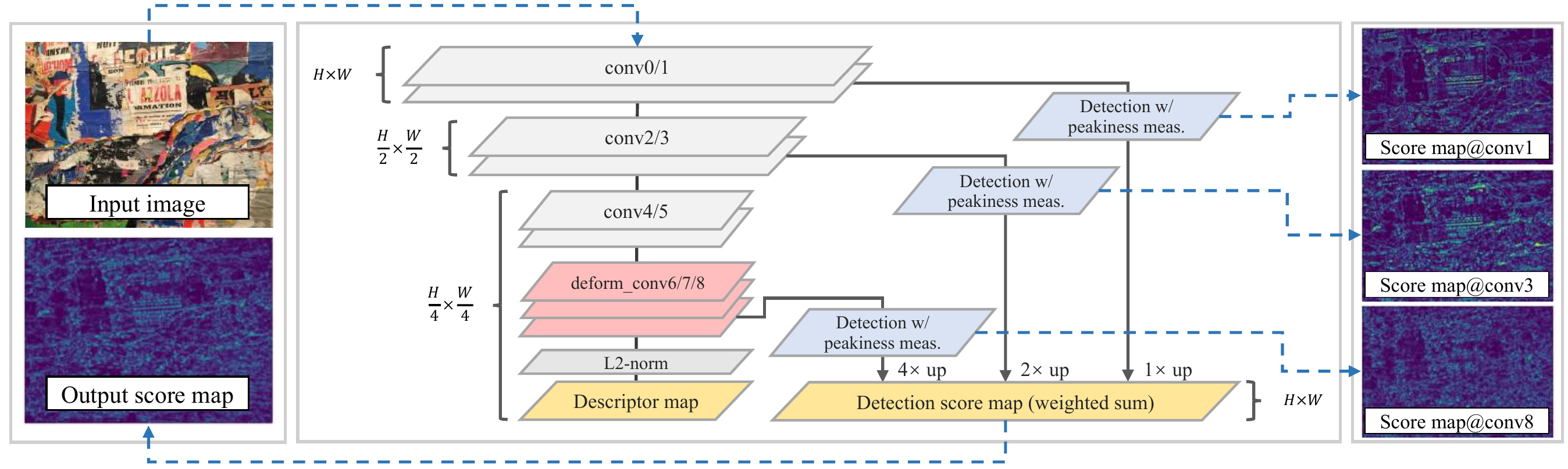}
    \caption{Network architecture, with the proposed  equipment of deformable convlutional network (DCN), multi-level detection (MulDet), and peakiness measurement for keypoint scoring.}
    \label{fig:net_arch}
\end{figure*}
The backbone architecture in this work is built upon 1) deformable convolutional networks (DCN)~\cite{dai2017deformable,zhu2019deformable} that predict and apply dense spatial transformation, and 2) D2-Net~\cite{dusmanu2019d2} that jointly learns keypoint detector and descriptor.

\smallskip\noindent\textbf{Deformable convolutional networks (DCN)~\cite{dai2017deformable,zhu2019deformable}} target to learn dynamic receptive filed to accommodate the ability of modelling geometric variations. Formally, given a regular grid $\mathcal{R}$ that samples values over the input feature maps $\mathbf{x}$, the output features $\mathbf{y}$ of a standard convolution for each spatial position $\mathbf{p}$ can be written as:
\begin{equation}
\label{equ:regular_conv}
\mathbf{y}(\mathbf{p})=\sum_{\mathbf{p}_n\in \mathcal{R}}\mathbf{w}(\mathbf{p}_n)\cdot \mathbf{x}(\mathbf{p} + \mathbf{p}_n).
\end{equation}
DCN augments the regular convolution by additionally learning both sampling offsets~\cite{dai2017deformable} $\{\bigtriangleup \mathbf{p}_n | n=1, ..., N\}$ and feature amplitudes~\cite{zhu2019deformable} $\{\bigtriangleup \mathbf{m}_n|n=1,...,N\}$, where $N=|\mathcal{R}|$, and rewrites  Eq.~\ref{equ:regular_conv} as:
\begin{equation}
\label{equ:deform_conv_v2}
\mathbf{y}(\mathbf{p})=\sum_{\mathbf{p}_n\in \mathcal{R}}\mathbf{w}(\mathbf{p}_n)\cdot \mathbf{x}(\mathbf{p} + \mathbf{p}_n + \bigtriangleup \mathbf{p}_n) \cdot \bigtriangleup \mathbf{m}_n.
\end{equation}
As the offset $\bigtriangleup \mathbf{p_n}$ is typically fractional, Eq.~\ref{equ:deform_conv_v2} is implemented via bilinear interpolation, while the feature amplitude $\bigtriangleup \mathbf{m}_n$ is limited to $(0, 1)$.  During training, the initial values of $\bigtriangleup \mathbf{p}_n$ and $\bigtriangleup \mathbf{m}_n$ are respectively set to $0$ and $0.5$, following the settings in~\cite{zhu2019deformable}. 


\smallskip\noindent\textbf{D2-Net}~\cite{dusmanu2019d2} proposes a \emph{describe-and-detect} strategy to jointly extract feature descriptions and detections. Over the last feature maps $\mathbf{y}\in\mathbb{R}^{H\times W \times C}$, D2-Net applies channel-wise L2-normalization to obtain dense feature descriptors, while the feature detections are derived from 1) the local score and 2) the channel-wise score. Specifically, for each location $(i, j)$ in $\mathbf{y}^c$ ($c=1, 2, ..., C$), the local score is obtained by:
\begin{equation}
\label{equ:local_max}
    \alpha^c_{ij} = \frac{\exp{(\mathbf{y}^c_{ij})}}{\sum_{{(i', j')}\in\mathcal{N}(i, j)} \exp{\mathbf{y}^c_{i'j'}}}, 
\end{equation}
where $\mathcal{N}(i, j)$ is neighboring pixels around $(i, j)$, e.g., 9 neighbours defined by a $3\times3$ kernel. Next, the channel-wise score is obtained by:
\begin{equation}
\label{equ:channel_max}
    \beta^c_{ij} = \mathbf{y}^c_{ij} / \max_t \mathbf{y}^t_{ij}.
\end{equation}
The final detection score is combined as:
\begin{equation}
\label{equ:d2net_score}
    s_{ij} = \max_t (\alpha^c_{ij} \beta^c_{ij}).
\end{equation}
The detection score will be later used as a weighting term in loss formulation (Sec.~\ref{sec:learning_framework}), and will allow for top-K selection of keypoints during testing.

\subsection{DCN with Geometric Constraints}

The original free-form DCN predicts local transformation of high degrees of freedom (DOF), e.g., $9\times 2$ offsets for a $3\times3$ kernel. On the one hand, it enables the potential to model complex deformation such as non-planarity, while on the other hand, it takes a risk of over-paramertizing the local shape, where simpler affine or perspective transformation are often considered to serve as a good approximation~\cite{mikolajczyk2002affine,moo2016learning,mishkin2018repeatability}. To find out what deformation is needed in our context, we compare three shape modellings via enforcing different geometric constraints in DCN, including 1) similarity, 2) affine and 3) homography. The shape properties of the investigated variants are summarized in Tab.~\ref{tab:dcn}.

\begin{table}[ht]
\centering
\resizebox{0.38\textwidth}{!}{ 
\begin{tabular}{ccc}
\Xhline{1pt}
\textbf{Variants}       & \textbf{Modeling Power}         & \textbf{DOF}    \\ \hline
\multicolumn{1}{l}{\textit{unconstrained}} & non-planarity          & $2k^2$ \\ \hdashline
\multicolumn{1}{l}{\textit{s.t. similarity}} & scale, rotation        & 2                     \\
\multicolumn{1}{l}{\textit{s.t. affine}}     & scale, rotation, shear & 4                     \\
\multicolumn{1}{l}{\textit{s.t. homography}} & perspective            & 6                     \\ \Xhline{1pt}
\end{tabular}
}
\caption{The shape properties of DCN variants, where DOF denotes the degrees of freedom and $k$ denotes the kernel size of convolution. Translation is omitted as is fixed for keypoints.}
\label{tab:dcn}
\end{table}

\smallskip\noindent\textbf{Affine-constrained DCN.} Traditionally, the local shape is often modelled by similarity transformation with estimates of rotation and scale~\cite{lowe2004distinctive,rublee2011orb}. In a learning framework such as~\cite{moo2016learning,ono2018lf}, this transformation is decomposed as:

\begin{equation}
\label{equ:sim_compos}
    \mathbf{S} = \lambda R(\theta) = \lambda
    \begin{pmatrix}
    \cos(\theta) & \sin(\theta) \\
    -\sin(\theta) & \cos(\theta)
    \end{pmatrix}.
\end{equation}
Moreover, a few works such as HesAff~\cite{mikolajczyk2002affine} further includes an estimate of shearing, which is cast as a learnable problem by AffNet~\cite{mishkin2018repeatability}. Here, we follow AffNet and decompose the affine transformation as:
\begin{align}
\begin{split}
\label{equ:aff_compos}
    \mathbf{A} &= \mathbf{S} A' = \lambda R(\theta) A' \\ 
    &=\lambda 
    \begin{pmatrix}
    \cos(\theta) & \sin(\theta) \\
    -\sin(\theta) & \cos(\theta)
    \end{pmatrix}
    \begin{pmatrix}
    a'_{11} & 0 \\
    a'_{21} & a'_{22}
    \end{pmatrix},
\end{split}
\end{align}
where $\det A'=1$. The network is implemented to predict one scalar for scaling ($\lambda$), another two for rotation ($\cos(\theta)$, $\sin(\theta))$, while the other three for shearing ($A'$).

\smallskip\noindent\textbf{Homography-constrained DCN.} Virtually, the local deformation can be better approximated by a homography (perspective) transformation $\mathbf{H}$, and we here adopt the Tensor Direct Linear Transform (Tensor DLT)~\cite{nguyen2018unsupervised} to solve the 4-point parameterization of $\mathbf{H}$ in a differentiable manner.

Formally, a linear system can be created that solves $\mathbf{M}\mathbf{h}=\mathbf{0}$, where $\mathbf{M}\in\mathbb{R}^{8\times 9}$ and $\mathbf{h}$ is a vector with 9 elements consisting of the entries of $\mathbf{H}$, and each correspondence provides two equations in $\mathbf{M}$. By enforcing the last element of $\mathbf{h}$ equals to 1~\cite{hartley2003multiple} and omitting the translation, we set $\mathbf{H}_{33}=1$ and $\mathbf{H}_{13}=\mathbf{H}_{23}=0$, then rewrite the above system of equations as $\hat{\mathbf{M}}_{(i)}\hat{\mathbf{h}}=\hat{\mathbf{b}}_{(i)}$, where $\hat{\mathbf{M}}_{(i)}\in\mathbb{R}^{2\times 6}$ and for each correspondence,
\begin{equation}
    \hat{\mathbf{M}}_{(i)}=
    \begin{bmatrix}
    0 & 0 & -u_i & -v_i & v_i'u_i & v_i'v_i \\
    u_i & v_i & 0 & 0 & -u_i'u_i & -u_i'v_i
    \end{bmatrix},
\end{equation}
$\hat{\mathbf{b}}_{(i)} = [-v'_i,u'_i]^T \in\mathbb{R}^{2\times 1}$ and $\hat{\mathbf{h}}$ consists of 6 elements from the first two columns of $\mathbf{H}$. By stacking the equations of 4 correspondences, we derive the final linear system:
\begin{equation}
    \hat{\mathbf{M}}\hat{\mathbf{h}}=\hat{\mathbf{b}}.
\end{equation}
Suppose that correspondence points are not collinear, $\hat{\mathbf{h}}$ can be then efficiently and uniquely solved by using the differentiable pseudo-inverse of $\hat{\mathbf{A}}$\footnote{Implemented via function \texttt{tf.matrix\_solve} in TensorFlow.}. In practice, we initialize 4 corner points at $\{(-1, -1), (1, -1), (1, 1), (-1, 1)\}$, and implement the network to predict 8 corresponding offsets lying in $(-1, 1)$ so as to avoid collinearity.

After forming the above transformation $\mathbf{T} \in \{ \mathbf{S}, \mathbf{A}, \mathbf{H}\}$, the offset values in Eq.~\ref{equ:deform_conv_v2} are now obtained by:
\begin{equation}
\label{equ:aff_offset}
    \bigtriangleup\mathbf{p}_n = \mathbf{T}\mathbf{p}_n - \mathbf{p}_n, \text{where}~\mathbf{p}_n\in\mathcal{R},
\end{equation}
so that geometry constraints are enforced in DCN. More implementation details can be found in the Appendix.

\subsection{Selective and Accurate Keypoint Detection}
\label{sec:detection}
\smallskip\noindent\textbf{Keypoint peakiness measurement.} As introduced in Sec.~\ref{sec:prer}, D2-Net scores a keypoint regarding both spatial and channel-wise responses. Among many possible metrics, D2-Net implements a \emph{ratio-to-max} (Eq.~\ref{equ:channel_max}) to evaluate channel-wise extremeness, whereas one possible limitation lies on that it only weakly relates to the actual distribution of all responses along the channel. 

To study this effect, we first trivially modify Eq.~\ref{equ:channel_max} with a channel-wise \texttt{softmax}, whereas this modification deteriorates the performance in our experiments. Instead, inspired by~\cite{revaud2019r2d2,zhang2018learning}, we propose to use \emph{peakiness} as a keypoint measurement in D2-Net, which rewrites Eq.~\ref{equ:channel_max} as:
\begin{equation}
    \beta^c_{ij} = \text{softplus}(\mathbf{y}^c_{ij} -  \frac{1}{C}\sum_{t}\mathbf{y}^t_{ij}),
\end{equation}
where \texttt{softplus} activates the peakiness to a positive value. To balance the scales of both scores, we also rewrites Eq.~\ref{equ:local_max} in the similar form:
\begin{equation}
    \alpha^c_{ij} = \text{softplus}(\mathbf{y}^c_{ij} - \frac{1}{|\mathcal{N}(i, j)|}{\sum_{{(i', j')}\in\mathcal{N}(i, j)} \mathbf{y}^c_{i'j'}}),
\end{equation}
and the two scores are again combined as in Eq.~\ref{equ:d2net_score}.

\begin{figure}
    \centering
    \includegraphics[width=0.48\textwidth]{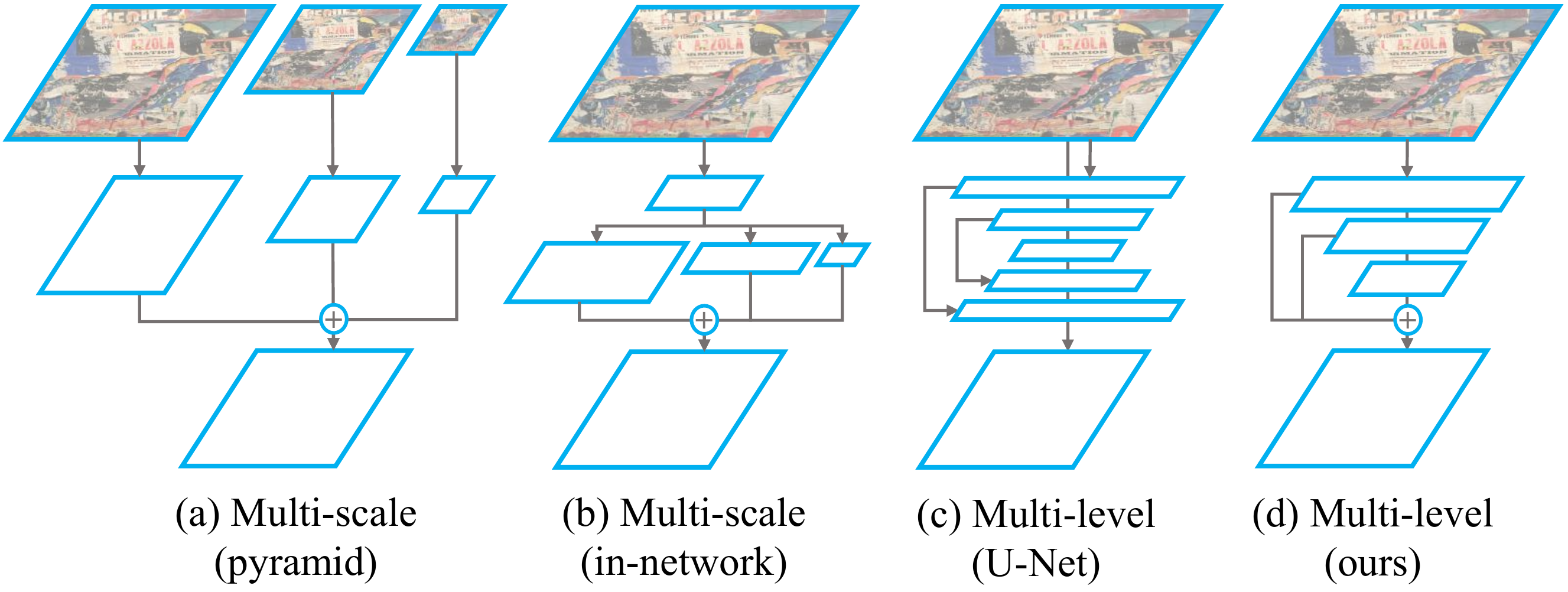}
    \caption{Different design choices to leverage feature hierarchy, shorten as variants of MulDet.}
    \label{fig:det}
\end{figure}

\smallskip\noindent\textbf{Multi-level keypoint detection (MulDet).} As aforementioned, one known limitation of D2-Net~\cite{dusmanu2019d2} is the lack of keypoint localization accuracy, since detections are obtained from low-resolution feature maps. There are multiple options to restore the spatial resolution, for instance, by learning an additional feature decoder (SuperPoint~\cite{detone2018superpoint}) or employing dilated convolutions (R2D2~\cite{revaud2019r2d2}). However, those methods either increase the number of learning parameters, or consume huge GPU memory or computation. Instead, we propose a simple yet effective solution without introducing extra learning weights, by leveraging the inherent pyramidal feature hierarchy of ConvNets and combining detections from multiple feature levels.

Specifically, given a feature hierarchy consisting of feature maps at different levels $\{\mathbf{y}^{(1)}, \mathbf{y}^{(2)}, ..., \mathbf{y}^{(l)}\}$ strided by $\{1, 2, ..., 2^{(l-1)}\}$, respectively, we apply the aforementioned detection at each level to get a set of score maps $\{\mathbf{s}^{(1)}, \mathbf{s}^{(2)}, ..., \mathbf{s}^{(l)}\}$. Next, each score map is upsampled to have the same spatial resolution as input image, and finally combined by taking the weighted sum:
\begin{equation}
\label{equ:muldet_score}
    \mathbf{\hat{s}}=\frac{1}{\sum_l w_{l}}\sum_l w_{l} \mathbf{s}^{(l)}.
\end{equation}
To better address the superiority of the proposed method, we implement 1) the multi-scale detection used in D2-Net~\cite{dusmanu2019d2} and R2D2~\cite{revaud2019r2d2} (Fig.~\ref{fig:det}{\color{red}a}) by constructing an image pyramid with multiple forward passes, 2) the in-network multi-scale prediction used in LF-Net~\cite{ono2018lf} (Fig.~\ref{fig:det}{\color{red}b}) by resizing the intermediate feature maps, and 3) the standard U-Net architecture~\cite{ronneberger2015u} (Fig.~\ref{fig:det}{\color{red}c}) that builds a feature decoder and skip connections from low-level feature maps.

The proposed multi-level detection (Fig.~\ref{fig:det}{\color{red}d}) is advantageous in three aspects. Firstly, it adopts implicit multi-scale detection that conforms to classical scale-space theory~\cite{lowe2004distinctive} by having different sizes of receptive field for localizing keypoints. Secondly, compared with U-Net architecture, it cheaply restores the spatial resolution without introducing extra learning weights to achieve pixel-wise accuracy. Thirdly, different from U-Net that directly fuses low-level and high-level features, it keeps the low-level features untouched, but fuses the \emph{detections} of multi-level semantics, which helps to better preserve the low-level structures such as corners or edges. Such representation representation is also similar to hypercolumns~\cite{hariharan2015hypercolumns}, which is used for object segmentation with fine-grained localization, The implementation details of above variants can be found in the Appendix.

\subsection{Learning Framework}
\label{sec:learning_framework}

\smallskip\noindent\textbf{Network architecture.} The network architecture is illustrated in Fig.~\ref{fig:net_arch}. To reduce computations, we replace the VGG backbone~\cite{simonyan2014very} used in D2-Net with more light-weight L2-Net~\cite{tian2017l2}. Similar to R2D2~\cite{revaud2019r2d2}, we further replace the last $8\times8$ convolution of L2-Net by three $3\times 3$ convolutions, resulting in feature maps of $128$ dimension and $1/4$ times resolution of the input. Finally, the last three convolutions, \texttt{conv6}, \texttt{conv7} and \texttt{conv8}, are substituted with DCN (Sec.~\ref{sec:prer}). Three levels, \texttt{conv1}, \texttt{conv3} and \texttt{conv8}, are selected to perform the proposed MulDet (Sec.~\ref{sec:detection}). The combination weights in Eq.~\ref{equ:muldet_score} are empirically set to $w_i={1,2,3}$, and the dilation rate to find neighboring pixels $\mathcal{N}(i, j)$ in Eq.~\ref{equ:local_max} is set to $3, 2, 1$, respectively, which we find to deliver best trade-offs to balance the attention on low-level and abstracted features. 

\smallskip\noindent\textbf{Loss design.} We identify a set of correspondences $\mathcal{C}$ for an image pair $(I, I')$ via densely warping $I$ to $I'$ regarding ground-truth depths and camera parameters. To derive the training loss for both detector and descriptor, we adopt the formulation in D2-Net~\cite{dusmanu2019d2}, written as:
\begin{equation}
    \mathcal{L}(I, I') = \frac{1}{|\mathcal{C}|}\sum_{c\in\mathcal{C}}\frac{\hat{s}_{c} \hat{s}'_{c}}{\sum_{q\in\mathcal{C}}\hat{s}_{q} \hat{s}'_{q}}\mathcal{M}(\mathbf{f}_c, \mathbf{f'}_c),
\end{equation}
where $\hat{s}_k$ and $\hat{s}'_k$ are combined detection scores in Eq.~\ref{equ:muldet_score} for image $I$ and $I'$, $\mathbf{f}_k$ and $\mathbf{f}'_k$ are their corresponding descriptors, and $\mathcal{M}(\cdot,\cdot)$ is the ranking loss for representation learning. Instead of using the hardest-triplet loss in D2-Net~\cite{dusmanu2019d2}, we adopt the hardest-contrastive form in FCGF~\cite{FCGF2019}, which we find guarantee better convergence when training from scratch and equipping DCN, written as:
\begin{align}
\begin{split}
  \mathcal{M}&(\mathbf{f}_c, \mathbf{f}'_c) = [D(\mathbf{f}_c,  \mathbf{f}'_c) - m_{p}]_+ + \\
    & [m_{n} - \min(\min_{k \neq c}D(\mathbf{f}_c, \mathbf{f}'_k), \min_{k \neq c}D(\mathbf{f}_k, \mathbf{f}'_c))]_+,
\end{split}
\end{align}
where $D(\cdot,\cdot)$ denotes the Euclidean distance measured between two descriptors, and $m_p,m_n$ are respectively set to $0.2, 1.0$ for positives and negatives. Similar to D2-Net~\cite{dusmanu2019d2}, a safe radius sized 3 is set to avoid taking spatially too close feature points as negatives.

\subsection{Implementations}
\label{sec:impl}
\smallskip\noindent\textbf{Training.} In contrast to D2-Net~\cite{dusmanu2019d2} which starts from an ImageNet pretrained model with only the last convolution fine-tuned, we train our model \emph{from scratch} with ground-truth cameras and depths obtained from GL3D~\cite{shen2018matchable} and~\cite{radenovic2016cnn} (the same data used in~\cite{luo2018geodesc,luo2019contextdesc}). The training consumes $800K$ image pairs sized $480\times 480$ and batched $2$. Learning gradients are computed for image pairs that have at least $32$ matches, while the maximum match number is limited to $512$. Each input image is standardized to have zero mean and unit norm, and independently applied with random photometric augmentation including brightness, contrast and blurriness. The SGD optimizer is used with momentum of 0.9, and the base learning rate is set to $0.1$.

Although an end-to-end learning with DCN is feasible, we find that a two-stage training yields better results in practice. Specifically, in the first round we train the model with \emph{all regular convolutions} for $400K$ iterations. In the second round, we tune \emph{only the DCNs} with the base learning rate divided by $10$ for another $400K$ iterations. Our implementation is made in TensorFlow with single NVIDIA RTX 2080Ti card, and the training finishes within $42$ hours.

\smallskip\noindent\textbf{Testing.} A non-maximum suppression (NMS) sized $3$ is applied to remove detections that are spatially too close. Similar to D2-Net, we postprocess the keypoints with the SIFT-like edge elimination (with threshold set to $10$) and sub-pixel refinement, the descriptors are then bilinearly interpolated at the refined locations. We select top-K keypoints regarding detection scores obtained in Eq.~\ref{equ:muldet_score}, and empirically discard those whose scores are lower than $0.50$.

\subsection{Post-CVPR Update}
\label{sec:post}
\begin{figure}[th]
	\centering
	\includegraphics[width=0.48\textwidth]{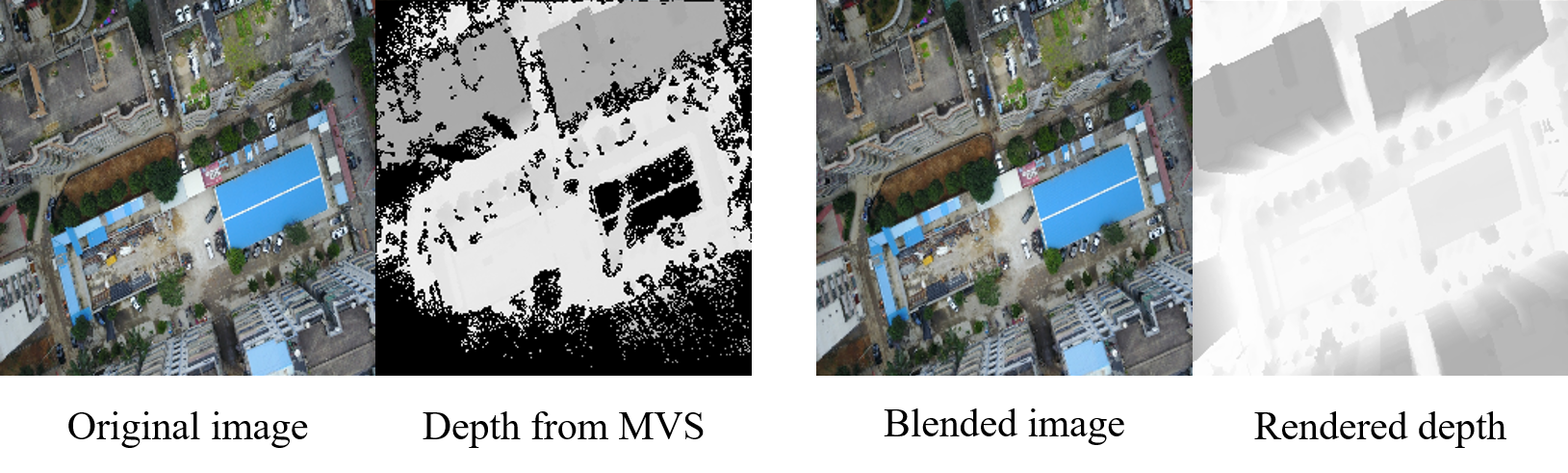}
	\caption{Training data is cleaned by generating rendered depths and blended images from 3D mesh models.}
	\label{fig:blended}
\end{figure}

\smallskip\noindent\textbf{Training data cleaning.} 
The above learning scheme requires dense warping via ground-truth cameras and depth maps for identifying ground-truth correspondences. However, both D2-Net and our method use depth maps that are produced by MVS algorithm (with proper post-processing), which may not be perfectly accurate or complete, probably resulting in noisy losses or raising difficulty in identifying larger number of correspondences. Inspired by BlendedMVS~\cite{yao2019blendedmvs}, we in addition generate rendered depth maps and blended images from 3D mesh models in GL3D~\cite{shen2018matchable}, so as to better align images and depths during training, as shown in Fig.~\ref{fig:blended}. We have combined BlendedMVS and GL3D together and made these data publicly available, further refer it to as BlendedMVG for solving general multi-view geometry problems.

\smallskip\noindent\textbf{Circle loss.} Using the hardest-contrastive loss in Sec.~\ref{sec:learning_framework} does not enjoy much benefit from the above data cleaning. Instead, we use the newly proposed circle loss~\cite{sun2020circle} to boost the performance. Empirically, we set the two hyparameters $m$ and $\gamma$ in circle loss to $0.1$ and $512$, respectively. In our experiments, we find that circle loss is robust to various parameter settings and guarantees better convergence.

\smallskip\noindent\textbf{Early stopping.} After applying above two updates, we find that the convergence becomes notably faster, whereas overfitting may occur especially during the second round of training (i.e., with DCNs). We thus conduct early stopping around $100K$ iterations when training with DCNs, instead of $400K$ as described in Sec.~\ref{sec:impl}.
 
\section{Experiments}
In the following sections, we evaluate our methods across several practical scenarios, including image matching, 3D reconstruction and visual localization. Further experiments on dense reconstruction and image retrieval can be found in the Appendix.
\subsection{Image Matching}
\label{sec:image_matching}
\smallskip\noindent\textbf{Datasets.}
First, we use the popular HPatches dataset~\cite{balntas2017hpatches}, which includes $116$ image sequences with ground-truth homography. Following D2-Net~\cite{dusmanu2019d2}, we exclude $8$ high-resolution sequences, leaving $52$ and $56$ sequences with illumination or viewpoint variations, respectively.

Though widely used, HPatches dataset exhibits only homography transformation, which may not comprehensively reflect the performance in real applications. Thus, we resort to the newly proposed FM-Bench~\cite{bian2019evaluation}, which comprises four datasets captured in practical scenarios: the TUM dataset~\cite{sturm2012benchmark} in indoor SLAM settings, the KITTI dataset~\cite{geiger2012we} in driving scenes, the Tanks and Temples dataset (T\&T)~\cite{knapitsch2017tanks} for wide-baseline reconstruction, and the Community Photo Collection (CPC)~\cite{wilson2014robust} for wild reconstruction from web images. For each datasets, $1000$ overlapping image pairs are randomly chosen for evaluation, with ground-truth fundamental matrix pre-computed.

\smallskip\noindent\textbf{Evaluation protocols.}
On HPatches dataset~\cite{balntas2017hpatches}, three standard metrics are used: 1) Keypoint repeatability (\emph{\%Rep.})\footnote{We initially did not apply a symmetric check when computing this metric, which is now fixed in the released code.}, a.k.a. the ratio of possible matches and the minimum number of keypoints in the shared view. 2) Matching score (\emph{\%M.S.}), a.k.a. the ratio of correct matches and the minimum number of keypoints in the shared view. 3) Mean matching accuracy (\emph{\%MMA}), a.k.a. the ratio of correct matches and possible matches. Here, a match is defined to correspond if the point distance is below some error threshold after homography wrapping, and a correct match is further required to be a mutual nearest neighbor during brute-force searching. For above metrics, we report their average scores over all image pairs in the dataset.

In terms of FM-Bench~\cite{bian2019evaluation}, a full matching pipeline including outlier rejection (e.g., ratio test~\cite{lowe2004distinctive}) and geometric verification (e.g., RANSAC) is performed, and the final pose recovery accuracy is evaluated. To determine the correctness of a pose estimate, FM-Bench uses ground-truth pose to generate a set of virtual correspondences, then measures the average of normalized symmetric epipolar distance regarding a pose estimate, and finally computes \emph{\%Recall} as the ratio of estimates where the distance error is below a certain threshold (0.05 by default). At correspondence level, FM-Bench also reports intermediate results such as the inlier ratio (\emph{\%Inlier/\%Inlier-m}) and correspondence number (\emph{\%Corr/\%Corr-m}) after/before RANSAC.

\begin{table}[th]
\resizebox{0.48\textwidth}{!}{ 
\begin{tabular}{llccc}
\Xhline{1pt}
\multicolumn{5}{c}{\textbf{HPatches dataset (error threshold {@} 3px)}}  \\ \hline
\multicolumn{1}{l|}{}                                   & \multicolumn{1}{l|}{\textit{Config.}}          & \textit{\%Rep.}        & \textit{\%M.S.}        & \textit{\%MMA} \\ \Xhline{1pt}
\multicolumn{1}{l|}{\multirow{3}{*}{\textbf{D2-Net}}}   & \multicolumn{1}{l|}{\textit{orig.}}                                &     \textbf{47.86}                &      23.58               &   43.00     \\
\multicolumn{1}{l|}{}                                   & \multicolumn{1}{l|}{\textit{our impl.}}                    &     43.34                 &         29.55             & 45.36  \\ \cdashline{2-5} 
\multicolumn{1}{l|}{}                                   & \multicolumn{1}{l|}{\textit{peakiness meas.}}                    &     46.24                  &         \textbf{32.27}             &  \textbf{48.54} \\ \hline\hline
\multicolumn{1}{l|}{\multirow{4}{*}{\textbf{+ MulDet}}}                                   & \multicolumn{1}{l|}{\textit{multi-scale (pyramid)}}       & 46.12 & 32.55 & 48.72       \\ 
\multicolumn{1}{l|}{} & \multicolumn{1}{l|}{\textit{multi-scale  (in-network)}}        & 45.17 & 31.74 & 47.94       \\
\multicolumn{1}{l|}{} & \multicolumn{1}{l|}{\textit{multi-level (U-Net)}}        & 75.35 & 40.12 & 66.42      \\
\cdashline{2-5} 
\multicolumn{1}{l|}{}                                   & \multicolumn{1}{l|}{\textit{multi-level (ours)}}       &  \textbf{77.37} & \textbf{42.99} &  \textbf{68.66}      \\ \hline\hline
\multicolumn{1}{l|}{}                    & \multicolumn{1}{l|}{\textit{s.t. similarity}} & 78.33 & 44.79 & 71.67       \\
\multicolumn{1}{l|}{\textbf{+ MulDet}} & \multicolumn{1}{l|}{\textit{s.t. affine}} & 78.49 & 45.35 & 71.80       \\
\multicolumn{1}{c|}{\multirow{2}{*}{\textbf{\&}}} & \multicolumn{1}{l|}{\textit{s.t. homography}} & 78.39 & 45.08 & 71.89      \\ \cdashline{2-5}
\multicolumn{1}{l|}{} & \multicolumn{1}{l|}{\textit{free-form, 1 layer}} &  78.27 & 45.12 &  71.08      \\
\multicolumn{1}{c|}{\textbf{DCN}} & \multicolumn{1}{l|}{\textit{free-form}} & 78.31 & \textbf{46.28} & 72.26       \\ 
\multicolumn{1}{l|}{} & \multicolumn{1}{l|}{\textit{free-form, multi-scale}} &  \textbf{86.03} & 39.37 & \textbf{72.64}       \\
\Xhline{1pt}
\end{tabular}
}
\caption{Ablation experiments of the proposed modifications, where \textbf{peakiness meas.} improves the detection scoring upon D2-Net, \textbf{+ MulDet} studies the effect of different feature fusion strategies, and \textbf{+ MulDet \& DCN} further compares the effect of different parameterzation of deformation.}
\label{tab:ablation}
\end{table}

\smallskip\noindent\textbf{Comparative methods.} We compare our methods with 1)  patch descriptors, including HardNet++~\cite{mishchuk2017working} with SIFT~\cite{lowe2004distinctive} detector (\emph{SIFT + HN++}), or plus a shape estimator HesAffNet~\cite{mishkin2018repeatability} (\emph{HAN + HN++}). Also, ContextDesc~\cite{luo2019contextdesc} with SIFT detector (\emph{SIFT + ContextDesc}). 2) Joint local feature learning approaches including SuperPoint~\cite{detone2018superpoint}, LF-Net~\cite{ono2018lf}, D2-Net (fine-tuned)~\cite{dusmanu2019d2} and more recent R2D2~\cite{revaud2019r2d2}. Unless otherwise specified, we report either results reported in original papers, or derived from authors' public implementations with default parameters. We limit the maximum numbers of features of our methods to 5K and 20K on HPatches dataset and FM-Bench, respectively.

On FM-Bench, both the mutual check and ratio test~\cite{lowe2004distinctive} are applied to reject outliers before RANSAC. A ratio at $0.8$ is used for all methods except for D2-Net and R2D2\footnote{We use $0.95$ for D2-Net as suggested in its original paper, and conduct a parameter searching for LF-Net, SuperPoint and R2D2, obtaining the ratio at $0.8$, $0.8$, and $0.9$, respectively, to achieve overall best performance.}.

\smallskip\noindent\textbf{Baseline.} To avoid overstatement, we first present our re-implementation of D2-Net (\emph{our impl.}) as the baseline. As mentioned in Sec.~\ref{sec:learning_framework} and Sec.~\ref{sec:impl}, the new baseline differs from the original D2-Net (\emph{orig.}) in three aspects: 1) Different backbone architecture (L2-Net~\cite{tian2017l2} with $128$-d output vs. VGG~\cite{simonyan2014very} with $512$-d output). 2) Different loss formulation (hardest-contrastive~\cite{FCGF2019} vs. hardest-triplet~\cite{dusmanu2019d2}). 3) Different training settings (trained from scratch vs. fine-tuned only the last convolution from a pre-trained model). As shown in Tab.~\ref{tab:ablation} and Tab.~\ref{tab:fmbench}, the new baseline outperforms original D2-Net in general, while being more parameter- and computation-efficient regarding model size.

\begin{table*}[th]
\resizebox{\textwidth}{!}{ 
\begin{tabular}{l|cccc|cccc}
\Xhline{1pt}
                      & \multicolumn{4}{c|}{\textbf{TUM}~\cite{sturm2012benchmark} (indoor SLAM settings)}                                                   & \multicolumn{4}{c}{\textbf{KITTI}~\cite{geiger2012we} (driving settings)}                                                  \\ \hline
\textit{Methods}      & \textit{\%Recall} & \textit{\%Inlier} & \textit{\#Inlier-m} & \textit{\#Corrs (-m)} & \textit{\%Recall} & \textit{\%Inlier} & \textit{\#Inlier-m} & \textit{\#Corrs (-m)} \\ \Xhline{1pt}
\textbf{SIFT}~\cite{lowe2004distinctive}         & 57.40             & 75.33             & 59.21               & 65 (316)              & 91.70             & 98.20             & 87.40               & 154 (525)             \\
\textbf{SIFT + HN++~\cite{mishchuk2017working}}    & 58.90             & 75.74             & 62.07               & 67 (315)              & 92.00             & 98.21             & 91.25               & 159 (535)             \\
\textbf{HAN + HN++~\cite{mishkin2018repeatability}}    & 51.70             & 75.70             & 62.06               & 101 (657)              & 90.40             & 98.09             & 90.64               & 233 (1182)             \\
\textbf{SIFT + ContextDesc}~\cite{luo2019contextdesc} & 59.70 & 75.53 & 62.61 & 69 (325) & \textbf{92.20} & 98.23 & 91.92 & 160 (541) \\ \hdashline
\textbf{LF-Net (MS)~\cite{ono2018lf}}       &   53.00                &    70.97               &       56.25              &         143 (851)             &     80.40              &       95.38           &    84.66                 &        202 (1045)               \\
\textbf{D2-Net (MS)~\cite{dusmanu2019d2}}       &   34.50                &    67.61               &       49.01              &        74 (1279)              &      71.40              &       94.26           &    73.25                 &       103 (1832)                \\
\textbf{SuperPoint~\cite{detone2018superpoint}}   &   45.80                &     72.79              &     64.06                &     39 (200)                  &       86.10            &       98.11            &      91.52               &      73 (392)                 \\
\textbf{R2D2 (MS)~\cite{revaud2019r2d2}}   &   57.70                &     73.70              &     61.53                &     260 (1912)                  &       78.80            &       97.53            &      86.49               &      278 (1804)                 \\
 \hdashline     
\textbf{D2-Net (our impl.)}         &     39.10              &     70.09              &       61.58              &        64 (337)              &     70.80         &         97.04        &  91.97   &  81 (683) \\ 
\textbf{ASLFeat (w/o peakiness meas.)}         &     53.30             &        74.96           &      68.29              &          116 (703)             &       89.60            &        98.47           &     95.36               &        223 (1376)               \\ 
\textbf{ASLFeat}         &     \textbf{60.20}              &        76.34           &         69.09            &          148 (739)             &        \textbf{92.20}          &         98.69          &      \textbf{96.25}               &         444 (1457)              \\
\textbf{ASLFeat (MS)}         &     59.90              &        \textbf{76.72}           &         \textbf{69.50}            &          258 (1332)             &        \textbf{92.20}          &         \textbf{98.76}          &      96.16               &         630 (2222)              \\
 \Xhline{1pt}

                      & \multicolumn{4}{c|}{\textbf{T\&T}~\cite{knapitsch2017tanks} (wide-baseline reconstruction)}                                                  & \multicolumn{4}{c}{\textbf{CPC}~\cite{wilson2014robust} (wild reconstruction from web images)}                                                    \\ \Xhline{1pt}
\textbf{SIFT}         & 70.00             & 75.20             & 53.25               & 85 (795)              & 29.20             & 67.14             & 48.07               & 60 (415)              \\
\textbf{SIFT + HN++}    & 79.90             & 81.05             & 63.61               & 96 (814)              & 40.30             & 76.73             & 62.30               & 69 (400)              \\
\textbf{HAN + HN++}    & 82.50             & 84.71             & 70.29               & 97 (920)              & 47.40             & 82.58             & 72.22               & 65 (405)             \\
\textbf{SIFT + ContextDesc} & 81.60 & 83.32 & 69.92 & 94 (728) & 41.80 & 84.01 & 72.21 & 61 (306) \\ \hdashline
\textbf{LF-Net (MS)}       &   57.40                &   66.62               &       60.57             &       54 (362)               &      19.40              &       44.27           &    44.35                 &      50 (114)                 \\
\textbf{D2-Net (MS)}       &     68.40              &       71.79           &       55.51              &      78 (2603)                 &       31.30            &      56.57            &      49.85              &        84 (1435)               \\
\textbf{SuperPoint}   &       81.80            &        83.87           &       70.89              &        52 (535)               &      40.50             &           75.28        &      64.68               &       31 (225)                \\ 
\textbf{R2D2 (MS)}   &        73.00           &       80.81            &                     65.31  &       84 (1462)               &      43.00             &      82.40            &                    67.28 &     91 (954)                  \\ \hdashline
\textbf{D2-Net (our impl.)}         &     83.20              &    84.19              &        75.32             &         74 (1009)            &          46.60    &  83.72               &  77.31  & 51 (464) \\ 
\textbf{ASLFeat (w/o peakiness meas.)}         &      86.30             &        84.71           &       77.84              &          171 (1775)             &       49.50            &        85.80           &      80.39               &        97 (780)               \\ 
\textbf{ASLFeat}         &      \textbf{89.90}             &        85.33           &       79.08              &          295 (2066)             &       51.50            &        87.98           &      82.24               &        165 (989)               \\ 
\textbf{ASLFeat (MS)}         &      88.70             &        \textbf{85.68}           &       \textbf{79.74}              &          327 (2465)             &       \textbf{54.40}            &        \textbf{89.33}           &      \textbf{82.76}               &        185 (1159)               \\ 
 \Xhline{1pt}
\end{tabular}
}
\caption{Evaluation results on FM-Bench~\cite{bian2019evaluation} for pair-wise image matching, where \emph{\#Recall} denotes the percentage of accurate pose estimates, \emph{\#Inlier} and \emph{\#Inlier-m}, \emph{\#Corrs} and \emph{\#Corrs-m} denote the inlier ratio and correspondence number after/before RANSAC.}
\label{tab:fmbench}
\vspace{-0.5em}
\end{table*}

\smallskip\noindent\textbf{Ablations on peakiness measurement.} We first adopt the peakiness measurement for more indicative keypoint scoring (Sec.~\ref{sec:detection}). As shown in Tab.~\ref{tab:ablation}, this modification (\emph{peakiness meas.}) notably improves the results regarding all evaluation metrics on HPatches dataset. This effect is validated on FM-Bench, which is shown to apply for all different scenarios as shown in Tab.~\ref{tab:fmbench} (\emph{ASLFeat w/o peakiness meas.}). Our later modifications will be thus based on this model.

\smallskip\noindent\textbf{Ablations on MulDet.} As shown in Tab.~\ref{tab:ablation}, applying multi-scale detection solely does not take obvious effect, as spatial accuracy is still lacking. Instead, adopting multi-level detection, with spatial resolution restored, remarkably boosts the performance, which conforms the necessity of pixel-level accuracy especially when small pixel error is tolerated. It is also note-worthy that, despite less learning weights and computation, the proposed multi-level detection outperforms the U-Net variant, addressing the low-level nature of this task where a better preservation of low-level features is beneficial. Although the proposed multi-level detection also includes feature fusion of difference scales, we find that combining a more explicit multi-scale (pyramid) detection (\emph{free-form, multi-scale}) is in particular advantagous in order to handle the scale changes. This combination will be denoted as ASLFeat (MS) in the following context.

\smallskip\noindent\textbf{Ablations on DCN.} As shown in Tab.~\ref{tab:ablation}, all investigated variants of DCN are valid and notably boost the performance. Among those designs, the free-form variant  slightly outperforms the constrained version, despite the fact that HPatches datasets exhibit only homography transformation. This confirms that modelling non-planarity is feasible and useful for local features, and we thus opt for the free-form DCN to better handle geometric variations. Besides, we also implement a single-layer DCN (\emph{free-form, 1 layer}) that replaces only the last regular convolution (i.e., \texttt{conv8} in Fig.~\ref{fig:net_arch}), showing that stacking more DCNs is beneficial and the shape estimation can be learned progressively. 

\begin{figure}[h]
    \centering
    \vspace{-0.5em}
    \includegraphics[width=0.48\textwidth]{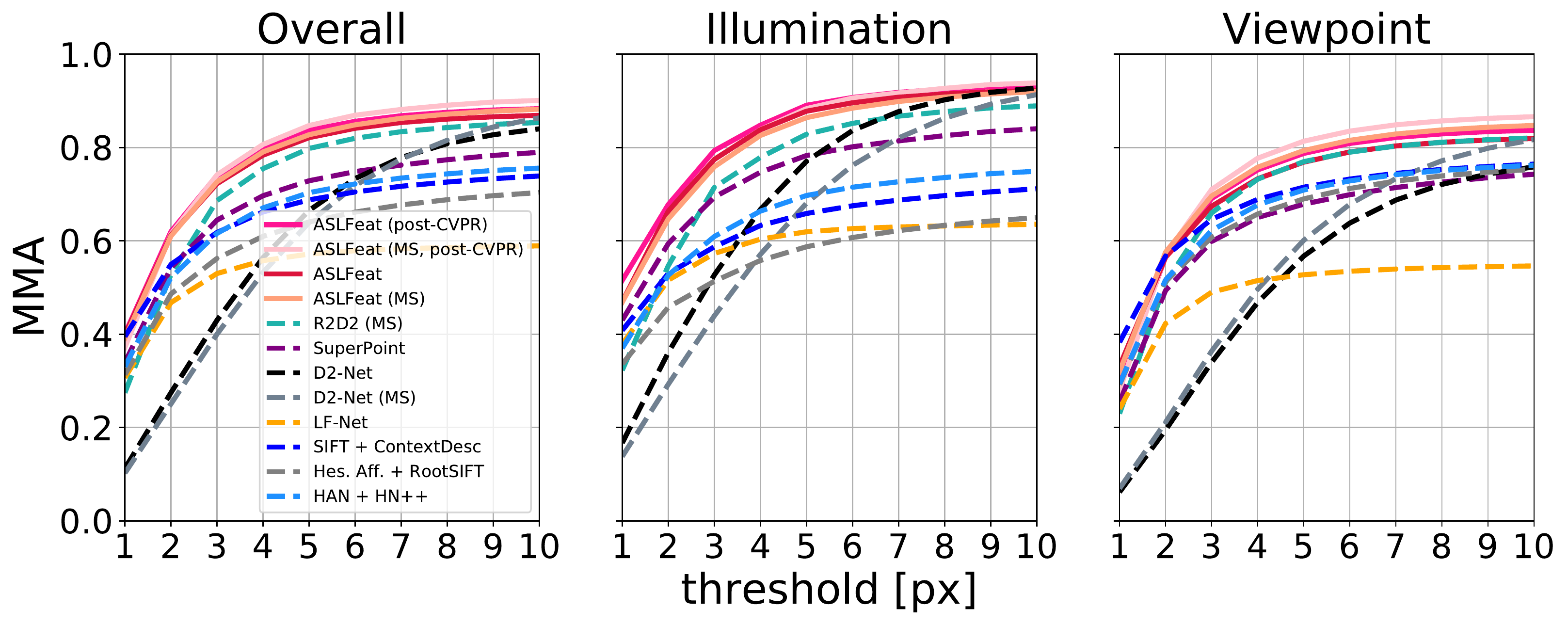}
    \caption{Comparisons on HPatches dataset~\cite{balntas2017hpatches} with mean matching accuracy (MMA) evaluated at different error thresholds, where ``MS" denotes that the multi-scale inference is enabled.}
    \label{fig:hseq_comp}
\end{figure}

\smallskip\noindent\textbf{Comparisons with other methods.} As illustrated in Fig.~\ref{fig:hseq_comp}, both ASLFeat and its multi-scale (MS) variant  achieve overall best results on HPatches dataset regarding both illumination and viewpoint variations at different error thresholds. Specifically, ASLFeat delivers remarkable improvements upon its backbone architecture, D2-Net, especially at low error thresholds, which in particular demonstrates that the keypoint localization error has been largely reduced. Besides, ASLFeat notably outperforms the more recent R2D2 ($72.64$ vs. $68.64$\footnote{By increasing  inference resolution, scale range, and enriching training data, R2D2 has achieved exciting results on this metric. Please visit their project page for details.} for MMA@3 overall), while being more computationally efficient by eschewing the use of dilated convolutions for restoring spatial resolution. By applying the post-CVPR updates in Sec.~\ref{sec:post}, ASLFeat (post-CVPR) and ASLFeat (MS, post-CVPR) achieve $74.02$ and $74.15$, respectively.

 In addition, as shown in Tab.~\ref{tab:fmbench} on FM-Bench, the ASLFeat remarkably outperforms other joint learning approaches. In particular, ASLFeat largely improves the state-of-the-art results on two MVS datasets: T\&T and CPC, of which the scenarios are consistent with the training data. It is also noteworthy that our methods generalize well to unseen scenarios: TUM (indoor scenes) and KITTI (driving scenes). As a common practice, adding more task-specific training data is expected to further boost the performance.

\smallskip\noindent\textbf{Visualizations.} We here present some sample detection results on FM-Bench in Fig.\ref{fig:det}, and more visualizations are provided in the Appendix.

\begin{figure}
    \centering
    \includegraphics[width=0.48\textwidth]{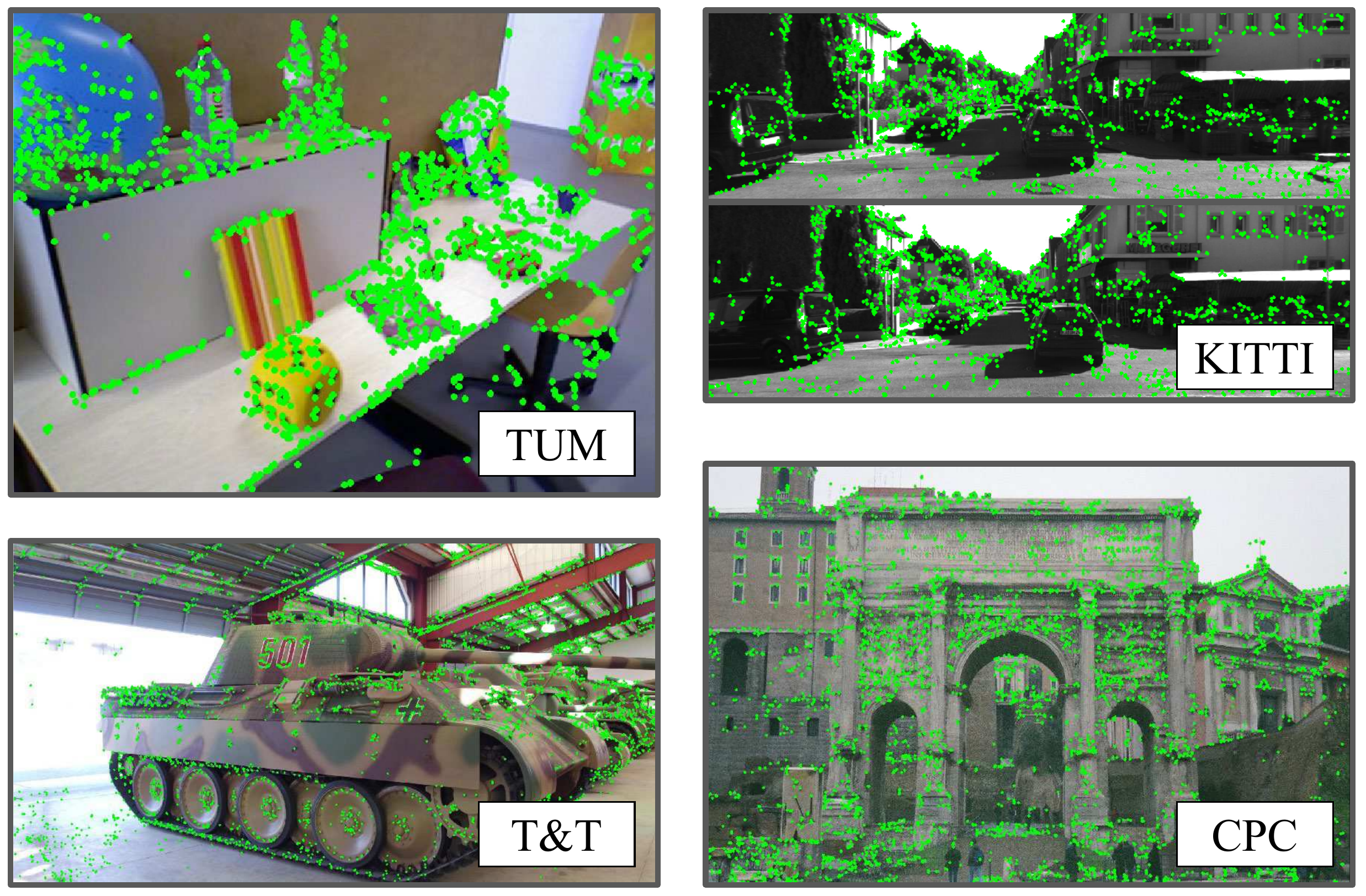}
    \caption{Sample detection results on FM-Bench~\cite{bian2019evaluation} with top-5000 keypoints displayed.}
    \label{fig:det}
\end{figure}

\subsection{3D Reconstruction}
\label{sec:3d_reconstruction}
\smallskip\noindent\textbf{Datasets.} We resort to ETH benchmark~\cite{schonberger2017comparative} to demonstrate the effect on 3D reconstruction tasks. Following~\cite{dusmanu2019d2}, we evaluate on three medium-scale datasets from~\cite{wilson2014robust}. 

\smallskip\noindent\textbf{Evaluation protocols.} We exhaustively match all image pairs for each dataset with both ratio test at $0.8$ and mutual check for outlier rejection, then run SfM and MVS algorithms by COLMAP~\cite{schonberger2016structure}. For sparse reconstruction, we report the number of registered images (\emph{\#Reg. Images}), the number of sparse points (\emph{\#Sparse Points}), average track length (\emph{Track Length}) and mean reprojection error (\emph{Reproj. Error}). For dense reconstruction, we report the number of dense points (\emph{\#Dense Points}). We limit the maximum number of features of ASLFeat to 20K.

\smallskip\noindent\textbf{Results.} As shown in Tab.~\ref{tab:eth_benchmark}, ASLFeat produces the most complete reconstructions regarding \emph{\#Reg. Images} and \emph{\#Dense Points}. Besides, ASLFeat results in \emph{Reproj. Error} that is on par with SuperPoint and smaller than D2-Net, which again validates the effect of the proposed MulDet for restoring spatial information. However, the reprojection error produced by hand-crafted keypoints (e.g., RootSIFT) is still notably smaller than all learning methods, which implies that future effort can be spent to further improve the keypoint localization in a learning framework.

\begin{table}[h]
\resizebox{0.5\textwidth}{!}{ 
\begin{tabular}{l|l|ccccc}
\Xhline{1pt}
\multirow{2}{*}{\textbf{Datasets}} & \multirow{2}{*}{\textbf{Methods}} & \textbf{\#Reg.} & \textbf{\# Sparse} & \textbf{Track}  & \textbf{Reproj.} & \textbf{\#Dense} \\
                                   &                                   & \textbf{Images} & \textbf{Points}    & \textbf{Length} & \textbf{Error}   & \textbf{Points}  \\ \hline
\textbf{Madrid}                    & \textit{RootSIFT}                 & 500             & 116K               & 6.32            & \textbf{0.60px}           & 1.82M            \\
\textbf{Metropolis}                & \textit{GeoDesc}                  & 495             & \textbf{144K}               & 5.97            & 0.65px           & 1.56M            \\ \cdashline{2-7}
\textbf{1344 images}                          & \textit{SuperPoint}                     &      438           &      29K              &      9.03         &      1.02px          &      1.55M            \\
\textbf{}               & \textit{D2-Net (MS)}              & 495             & 144K               & 6.39            & 1.35px           & 1.46M            \\  \cdashline{2-7}
\textbf{}                          & \textit{ASLFeat}                     &       613          &    96K                &     8.76            &     0.90px             &    \textbf{2.00M}              \\ 
\textbf{}                          & \textit{ASLFeat (MS)}                     &       
	\textbf{649}          &    129K                &        \textbf{9.56}         &      0.95px           &     1.92M              \\
\hline
\textbf{Gendarmen-}                & \textit{RootSIFT}                 & 1035            & 338K               & 5.52            & \textbf{0.69px}           & \textbf{4.23M}            \\
\textbf{markt}                     & \textit{GeoDesc}                  & 1004            & \textbf{441K}               & 5.14            & 0.73px           & 3.88M            \\ \cdashline{2-7}
\textbf{1463 images}                          & \textit{SuperPoint}                     &      967           &         93K           &      7.22          &    1.03px             &    3.81M              \\
\textbf{}               & \textit{D2-Net (MS)}              & 965             & 310K               & 5.55            & 1.28px           & 3.15M            \\ \cdashline{2-7}
\textbf{}                          & \textit{ASLFeat}                     &     1040            &      221K              &     8.72           &     1.00px             &     4.01M             \\ 
\textbf{}                          & \textit{ASLFeat (MS)}                     &       \textbf{1061}          &        320K            &        \textbf{8.98}         &     1.05px            &     4.00M              \\
\hline
\textbf{Tower of}                  & \textit{RootSIFT}                 & 804             & 239K               & 7.76            & \textbf{0.61px}           & 3.05M            \\
\textbf{London}                    & \textit{GeoDesc}                  & 776             & \textbf{341K}               & 6.71            & 0.63px           & 2.73M            \\ \cdashline{2-7}
\textbf{1576 images}                          & \textit{SuperPoint}                     &     681            &       52K             &      8.67          &     0.96px             &      2.77M            \\
\textbf{}               & \textit{D2-Net (MS)}              & 708             & 287K               & 5.20            & 1.34px           & 2.86M            \\ \cdashline{2-7}
\textbf{}                          & \textit{ASLFeat}                     &       821          &      222K              &        12.52         &      0.92px            &     3.06M              \\ 
\textbf{}                          & \textit{ASLFeat (MS)}                     &       \textbf{846}          &   252K                 &        \textbf{13.16}         &      0.95px           &     \textbf{3.08M}              \\
\Xhline{1pt}
\end{tabular}
}
\caption{Evaluation results on ETH benchmark~\cite{schonberger2017comparative} for 3D reconstruction.}
\label{tab:eth_benchmark}
\end{table}

\subsection{Visual Localization}

\smallskip\noindent\textbf{Datasets.} We resort to Aachen Day-Night dataset~\cite{sattler2012image} to demonstrate the effect on visual localization tasks, where the key challenge lies on matching images with extreme day-night changes for $98$ queries.

\smallskip\noindent\textbf{Evaluation protocols.} We use the evaluation pipeline provided in \emph{The Visual Localization Benchmark\footnote{https://www.visuallocalization.net/}}, which takes custom features as input, then relies on COLMAP~\cite{schonberger2016structure} for image registration, and finally generates the percentages of successfully localized images within three error tolerances (0.5m, 2$^\circ$) / (1m, 5$^\circ$) / (5m, 10$^\circ$). The maximum feature number of our methods are limited to 20K.

\smallskip\noindent\textbf{Results.} As shown in Tab.~\ref{tab:aachen}, although only mediocre results are obtained in previous evaluations, D2-Net performs surprisingly well under challenging illumination variations. This can be probably ascribed to the superior robustness of low-level features pre-trained on ImageNet. On the other hand, our method outperforms the plain implementation of R2D2, while a specialized R2D2 model (\emph{R2D2 (fine-tuned)}) achieves the state-of-the-art results with doubled model size, training on day image from Aachen dataset and using photo-realistic style transfer to generate night images.

\begin{table}[th]
\centering
\resizebox{0.48\textwidth}{!}{ 
\begin{tabular}{lcc|ccc}
\Xhline{1pt}
\multicolumn{1}{c}{\textbf{Methods}} & \textbf{\#Features} & \textbf{Dim} & \textbf{0.5m, 2$^\circ$} & \textbf{1m, 5$^\circ$} & \textbf{5m, 10$^\circ$} \\ \hline
\textit{RootSIFT}                    & 11K                 & 128          & 33.7             & 52.0           & 65.3            \\
\textit{HAN + HN++}                      & 11K                 & 128          & 37.8             & 54.1           & 75.5            \\
\textit{SIFT + ContextDesc}                 & 11K                 & 128          & 40.8             & 55.1           & 80.6            \\ \hdashline
\textit{SuperPoint}                  & 7K                  & 256          & 42.8             & 57.1           & 75.5            \\
\textit{D2-Net (MS)}                      & 19K                 & 512          & 44.9             & 64.3           & \textbf{88.8}            \\
\textit{R2D2 (MS)}                        & 10K                 & 128          & 43.9             & 61.2           & 77.6            \\
\textit{R2D2 (MS, fine-tuned)}                & 10K                 & 128          & 45.9             & \textbf{66.3}           & \textbf{88.8}            \\ \hdashline
\textit{D2-Net (our impl.)}                        & 10K                 & 128          & 40.8             & 59.2           & 77.6            \\
\textit{ASLFeat}                        & 10K                 & 128          & 45.9             & 64.3           & 86.7            \\ 
\textit{ASLFeat (MS)}                        & 10K                 & 128          & 44.9             & 64.3           & 85.7            \\  \hdashline
\textit{ASLFeat (post-CVPR)}                & 10K                 & 128          & \textbf{46.9}             & 65.3           & \textbf{88.8}            \\
\Xhline{1pt}
\end{tabular}
}
\caption{Evaluation results on Aachen Day-Night dataset~\cite{sattler2012image} for visual localization.}
\label{tab:aachen}
\end{table}
\vspace{-1.em}

\section{Conclusions}
In this paper, we have used D2-Net as the backbone architecture to jointly learn the local feature detector and descriptor. Three light-weight yet effective modifications have been proposed that drastically boost the performance in two aspects: the ability to model the local shape for stronger geometric invariance, and the ability to localize keypoints accurately for solving robust camera geometry. We have conducted extensive experiments to study the effect of each modification, and demonstrated the superiority and practicability of our methods across various applications.

\smallskip\noindent\textbf{Acknowledgments.} This work is supported by Hong Kong RGC GRF 16206819, 16203518 and T22-603/15N.

\clearpage
{\small
	\bibliographystyle{ieee}
	\bibliography{egbib}
}

\clearpage
\section*{A. Supplementary Appendix}

\subsection*{A.1 Implementation Details of DCN}
Since no native implementation of modulated DCN~\cite{zhu2019deformable} is currently available in TensorFlow, we implement DCN of our own. To reduce the number of learning weights, only one set of deformation parameters are predicted and then applied along all channels, similar to the setting when \texttt{num\_deformable\_groups=1} in PyTorch implementation\footnote{https://github.com/chengdazhi/Deformable-Convolution-V2-PyTorch/blob/master/modules/deform\_conv.py}.

To enforce DCN with affine constraints, we follow the implementation of AffNet~\cite{mishkin2018repeatability}, and construct a network to predict one bounded scalar to model the scaling factor in Eq.~\ref{equ:aff_compos}, formulated as:
\begin{equation}
    \lambda(x) = \exp{(\tanh{(x)})}.
\end{equation}
To model the rotation, two scalars are predicted as scaled cosine and sine, which are then used to compute an angle by taking:
\begin{equation}
    \theta(x, y) = \text{arctan2}{(x, y)}.
\end{equation}
To compose the affine shape matrix $A'$, we implement the network to predict the residual shape, and enforce $\det A'=1$ by:

\begin{equation}
A'=
    \begin{pmatrix}
    |1 + a''_{11}| & 0 \\
    a''_{21} & |1 + a''_{22}|
    \end{pmatrix}
    \begin{pmatrix}
    \frac{1}{|(1 + a''_{11})(1 + a''_{22})|} & 0 \\
    0 & 1
    \end{pmatrix},
\end{equation}
where $a''_{11}, a''_{21}$ and $a''_{22}$ lie in range $(-1, 1)$ through an $\tanh$ activation. In contrast to the observation in AffNet~\cite{mishkin2018repeatability}, we do not suffer degeneration when joint learning all affine parameters in DCN. 

In this paper, we have concluded that the free-form DCN is a preferable choice than other deformation parameterziation subject to geometric constraints, in the context of local feature learning. However, as shown in Tab.~\ref{tab:ablation}, this difference is not that obvious. We ascribe this phenomenon to the lack of meaningful supervision for learning complex deformation. As also discussed in AffNet~\cite{mishkin2018repeatability}, a specialized loss may be needed to guide the local shape estimation, whereas in our current implementation, the same loss is used in both local feature learning and deformation learning (we have tried the loss in AffNet~\cite{mishkin2018repeatability}, whereas no consistent improvement has been observed). In the future, we will further explore this direction in order to better release the potential of DCN.

\subsection*{A.2 Implementation Details of MulDet}

To implement the \emph{multi-scale (pyramid)} variant, we follow D2-Net~\cite{dusmanu2019d2} and R2D2~\cite{revaud2019r2d2}, and feed an image pyramid to the network, which is constructed from the input image sized up to $2048$, and  downsampled by $\sqrt{2}$ and blurred by a Gaussian kernel factored $0.8$ for each scale, until the longest side is smaller than 128 pixels. In each scale, a set of keypoints are identified whose scores are above some threshold, e.g., $0.5$, and the final top-K keypoints are selected from the keypoints combined from all scales. The inference time will be doubled when enabling this configuration.

To implement the \emph{multi-scale (in-network)} variant, we follow LF-Net~\cite{ono2018lf}, and resize the feature maps from the last convolution, i.e., \texttt{conv8}, into multiple scales. Specifically, the resizing is repeated for $N$ times, resulting scales from $1 / R$ to $R$, where $N=5$ and $R=\sqrt{2}$. Each corresponding score map is generated as Eq.~\ref{equ:d2net_score}, then the final scale-space score map is obtained by merging all the score maps via weighted-summation, where the weight is computed from a softmax function. Since DCN has already handled the in-network scale invariance, we did not find this variant useful when combining with our methods.

To implement the \emph{multi-level (U-Net)} variant, we build skip connections from two levels, i.e., \texttt{conv1} and \texttt{conv3}, and fuse different levels via feature concatenation. The same training scheme is applied as in the main paper, except that the keypoints are now derived from high-resolution feature maps.
\subsection*{A.3 Additional Experiments}
\smallskip\noindent\textbf{Evaluation on dense reconstruction.} In Sec.~\ref{sec:image_matching}, we have used T\&T dataset~\cite{knapitsch2017tanks} to evaluate the performance in two-view image matching. Here, we resort to its original evaluation protocols defined for evaluation dense reconstruction, and integrate ASLFeat into a dense reconstruction pipeline of our own to further demonstrate its superiority.

Specifically, we use the training set of T\&T, including $7$ scans with ground-truth scanned models, and use \emph{F-score} defined in~\cite{knapitsch2017tanks} to jointly measure the reconstruction accuracy (precision) and reconstruction completeness (recall). For comparison, we choose RootSIFT~\cite{arandjelovic2012three}, GeoDesc~\cite{luo2018geodesc} with SIFT detector~\cite{lowe2004distinctive}, and sample the features to 10K for each method. Next, we apply the same matching strategy (mutual check plus a ratio test at 0.8), SfM and dense algorithms to obtain the final dense point clouds. 
As shown in Tab.~\ref{tab:tt_supp}, ASLFeat delivers consistent improvements on dense reconstruction. Since T\&T exhibits less scale difference, ASLFeat without the multi-scale detection yields overall best results.

\begin{table}[ht]
	\centering
	\resizebox{0.48\textwidth}{!}{ 
		\begin{tabular}{l|cccc}
			\Xhline{1pt}
			\multicolumn{1}{c|}{\textbf{Methods}} & \textbf{RootSIFT~\cite{arandjelovic2012three}} & \textbf{GeoDesc~\cite{luo2018geodesc}} & \textbf{ASLFeat} & \textbf{ASLFeat (MS)} \\ \hline
			\textit{Barn}                         &      46.27             & 50.08            & \textbf{55.54} & 50.27        \\
			\textit{Caterpillar}                  &      50.72             & 48.87            & \textbf{51.70} & 48.88         \\
			\textit{Church}                       &       42.73            & \textbf{42.93}            & 42.66 & 37.82         \\
			\textit{Courthouse}                   &       43.11            & 43.96            & 44.41 & \textbf{50.39}         \\
			\textit{Ignatius}                     &       66.91            & 64.45            & \textbf{67.77} & 63.30         \\
			\textit{Meetingroom}                  &      19.89             & 20.39            & \textbf{26.59} & 25.39         \\
			\textit{Truck}                        &        67.67           & 67.86            & 70.43 & \textbf{71.31}         \\ \hdashline
			\textit{Mean}                         &       48.19            & 48.36            & \textbf{51.30} & 49.62         \\ \Xhline{1pt}
		\end{tabular}
	}
\caption{Evaluation results on T\&T dataset~\cite{knapitsch2017tanks} for dense reconstruction. The \emph{F-score} is reported to quantify both the reconstruction accuracy and reconstruction completeness.}
\label{tab:tt_supp}
\end{table}

\smallskip\noindent\textbf{Application on image retrieval.} We use an open-source implementation of VocabTree\footnote{https://github.com/hlzz/libvot}~\cite{shen2016graph} for evaluating image retrieval performance on the popular Oxford buildings~\cite{philbin2007object} and Paris dataset~\cite{philbin2008lost}. For clarity, we do not apply advanced post-processing (e.g., query expansion) or re-ranking methods (e.g., spatial verification), and report the mean average precision (mAP) for all comparative methods. For fair comparison, we sample the top-10K keypoints for each method to build the vocabulary tree. As shown in Tab.~\ref{tab:oxford_supp}, the proposed feature also performs well in this task, which further extends its usability in real applications.

\begin{table}[th]
	\centering
	\resizebox{0.48\textwidth}{!}{ 
	\begin{tabular}{c|ccccc}
		\Xhline{1pt}
		\textbf{} & \textbf{RootSIFT}~\cite{arandjelovic2012three} & \textbf{GeoDesc}~\cite{luo2018geodesc} & \textbf{ContextDesc}~\cite{luo2019contextdesc} & \textbf{ASLFeat}  & \textbf{ASLFeat (MS)}\\ \hline
		\emph{Oxford5k}     & 44.94    & 51.77   & 65.31       & 67.01 & \textbf{73.19}\\ 
		\emph{Paris6k}     &  45.83  & 48.15   & 60.79      & 58.01 & \textbf{64.96}\\
		\Xhline{1pt}
	\end{tabular}
}
	\caption{Evaluation results on Oxford buildings~\cite{philbin2007object} and Paris dataset~\cite{philbin2008lost} for image retrieval. The mean average precision (mAP) is reported.} 
	\label{tab:oxford_supp}
\end{table}

\smallskip\noindent\textbf{Integration with a learned matcher.}
In contrast to R2D2~\cite{revaud2019r2d2} which strengthens the model with additional task-specific training data and data augmentation by style transfer, we explore the usability of equipping a learnable matcher to reject outlier matches for improving the recovery of camera poses. Specifically, we resort to the recent OANet~\cite{zhang2019learning}, and train a matcher using the authors' public implementation\footnote{https://github.com/zjhthu/OANet.git} with ASLFeat. Finally, we integrate the resulting matcher into the evaluation pipeline of Aachen Day-Night dataset~\cite{sattler2012image}. 
As shown in Tab.~\ref{tab:aachen_supp}, this integration (\emph{ASLFeat + OANet}) further boosts the localization results.

\begin{table}[th]
	\centering
	\resizebox{0.48\textwidth}{!}{ 
		\begin{tabular}{l|ccc}
			\Xhline{1pt}
			\multicolumn{1}{l|}{\textbf{Methods}} & \textbf{0.5m, 2$^\circ$} & \textbf{1m, 5$^\circ$} & \textbf{5m, 10$^\circ$} \\ \hline
			\textit{ASLFeat}                    & 45.9             & 64.3           & 86.7            \\ 
			\textit{ASLFeat + OANet~\cite{zhang2019learning}} & \textbf{48.0} & \textbf{67.3} & \textbf{88.8} \\ \hdashline
			\textit{ASLFeat (MS)}                    & 44.9             & 64.3           & 85.7            \\ 
			\textit{ASLFeat (MS) + OANet} &  45.9 & \textbf{67.3} & 87.8 \\ 
			\Xhline{1pt}
		\end{tabular}
	}
	\caption{Evaluation results on Aachen Day-Night dataset~\cite{sattler2012image} for visual localization.}
	\label{tab:aachen_supp}
\end{table}

\subsection*{A.4 Discussions}
\smallskip\noindent\textbf{End-to-end learning with DCN.} As mentioned in Sec.~\ref{sec:impl}, we find that a two-stage training for deformation parameters yields better results, i.e., 72.64 for MMA@3 on HPatches dataset (Tab.~\ref{tab:ablation}), while an end-to-end training results in 70.45. Although we have tried different training strategies, e.g., dividing the learning rate of deformation parameters by 10 during end-to-end training, none of them have shown better results than the simple separate training. It is still under exploration whether an end-to-end learning will benefit more to this learning process. 

\begin{figure}[h]
    \centering
    \includegraphics[width=0.48\textwidth]{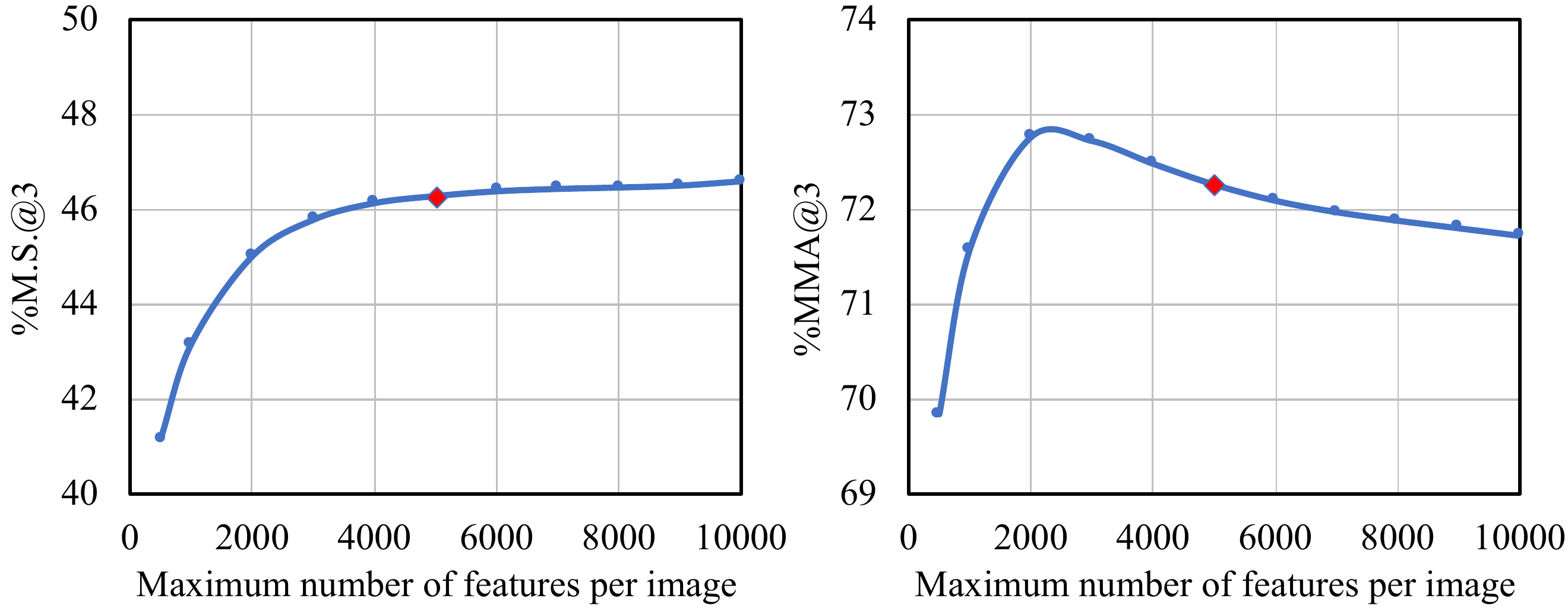}
    \caption{Matching score (M.S.) and mean matching accuracy (MMA) at an error threshold of 3px regarding different maximum numbers of features. We report the results at 5K features (marked in red) for our methods.}
    \label{fig:vary_feat_num}
\end{figure}

\smallskip\noindent\textbf{Performance regarding different feature number.}
In Fig.~\ref{fig:vary_feat_num}, we again use HPathces dataset~\cite{balntas2017hpatches}, and plot the performance change (M.S. and MMA) when limiting different maximum numbers of features.

\subsection*{A.5 More Visualizations}
\begin{figure*}[h]
	\centering
	\includegraphics[width=\textwidth]{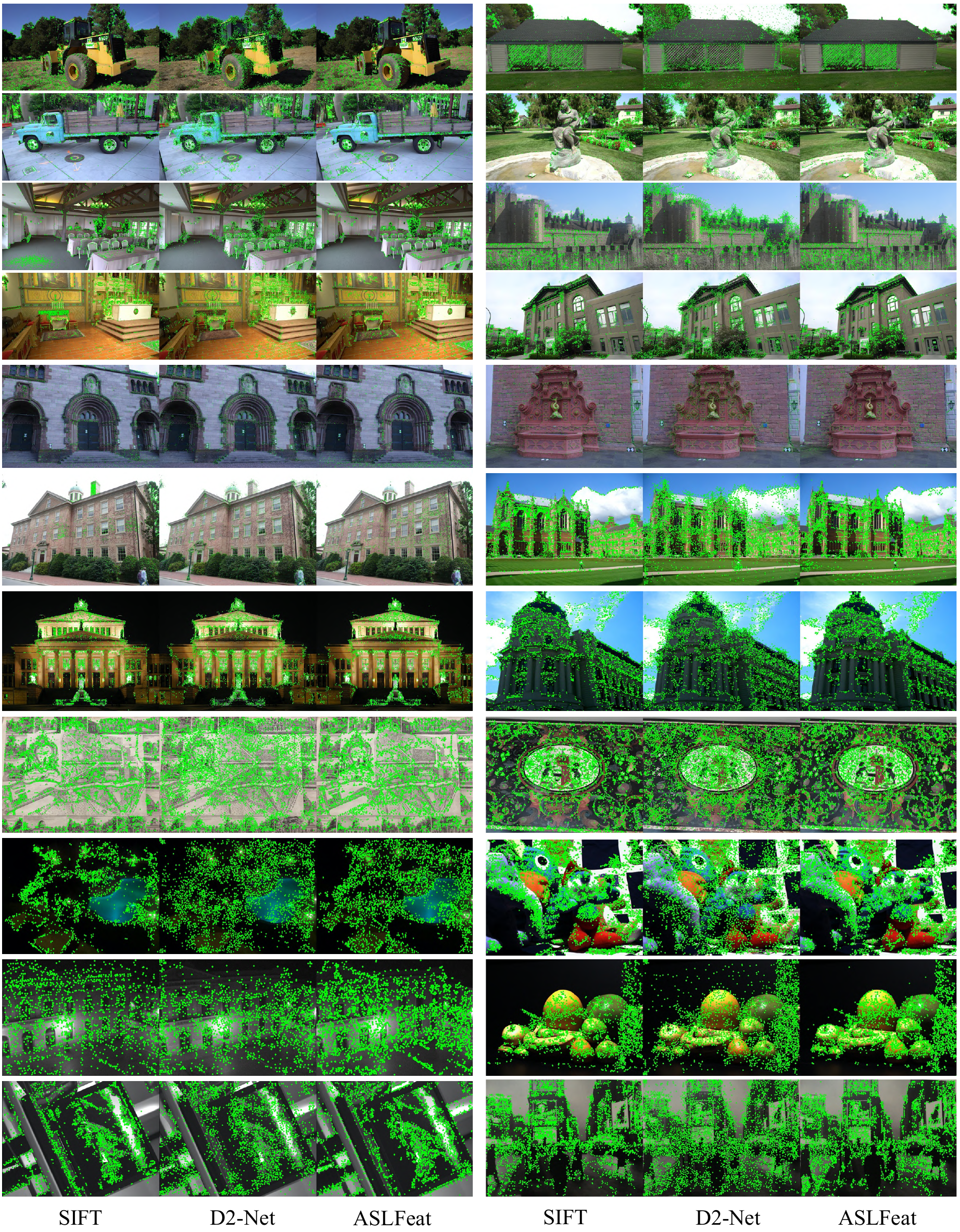}
	\caption{Comparisons of keypoints from different methods.}
	\label{fig:vis_supp}
\end{figure*}

We provide visualizations in Fig.~\ref{fig:vis_supp} for comparing the keypoints from different local features, including SIFT, D2-Net and the proposed method.

\end{document}